%% file: ACM2026/main.tex
\documentclass[manuscript,nonacm]{acmart}

\usepackage{multirow}
\usepackage{subcaption}
\usepackage{lineno}
\begin{document}

\title{\textsc{Argus}: Seeing the Influence of Narrative Features on Persuasion in Argumentative Texts}

\author{Sara Nabhani}
\email{s.nabhani@rug.nl}
\orcid{0009-0003-0599-5756}
\author{Federico Pianzola}
\email{f.pianzola@rug.nl}
\orcid{0000-0001-6634-121X}
\author{Khalid Al-Khatib}
\email{khalid.alkhatib@rug.nl}
\orcid{0009-0006-7255-5349}
\author{Malvina Nissim}
\email{m.nissim@rug.nl}
\orcid{0000-0001-5289-0971}
\affiliation{
  \institution{University of Groningen}
  \city{Groningen}
  \country{Netherlands}
}

\renewcommand{\shortauthors}{Nabhani et al.}

\begin{abstract}
\textit{Can narratives make arguments more persuasive? And to this end, which narrative features matter most?} Although stories are often seen as powerful tools for persuasion, their specific role in online, unstructured argumentation remains underexplored. To address this gap, we present \textsc{Argus}, a framework for studying the impact of narration on persuasion in argumentative discourse. \textsc{Argus} introduces a new ChangeMyView corpus annotated for story presence and six key narrative features, integrating insights from two established theoretical frameworks that capture both textual narrative features and their effects on recipients. Leveraging both encoder-based classifiers and zero-shot large language models (LLMs), \textsc{Argus} identifies stories and narrative features and applies them at scale to examine how different narrative dimensions influence persuasion success in online argumentation. 
\end{abstract}

\maketitle

\input{ACM2026/sections/introduction.tex}
\input{ACM2026/sections/related_work.tex}
\input{ACM2026/sections/data.tex}

\input{ACM2026/sections/modeling.tex}

\input{ACM2026/sections/analysis.tex}
\input{ACM2026/sections/conclusion.tex}
% \begin{acks}

% \end{acks}

\bibliographystyle{ACM2026/ACM-Reference-Format.bst}
\bibliography{ACM2026/custom.bib}

\appendix
\input{ACM2026/sections/appendix.tex}

\end{document}

%% file: ACM2026/sections/introduction.tex
\section{Introduction}
Argumentation is a central mode of communication in online discourse, where individuals justify their beliefs, challenge opposing viewpoints, and attempt to persuade others. Computational argumentation has made substantial progress in modeling these interactions, particularly in identifying strategies that differentiate persuasive from non-persuasive texts~\cite{saenger:2024}. In parallel, narrative has been extensively studied across disciplines for its rhetorical power, with evidence of its capacity to enhance persuasion through mechanisms such as emotional engagement and cognitive transportation~\cite{GREEN20241}.

Despite their importance, narratives in online argumentative discourse have received little attention. Most research draws only a broad distinction between stories and non-stories, without examining the specific elements that make up a narrative. However, these finer elements may each contribute to persuasiveness in distinct ways.

Grounded in two well-established narrativity theories \cite{herman2009basic, sternberg:2003, Pianzola2018Looking}, we introduce \textsc{Argus}, a framework for analyzing the role of narratives in online argumentative discourse. \textsc{Argus} combines a new corpus of annotated texts from the \textit{ChangeMyView} forum\footnote{\url{https://www.reddit.com/r/changemyview/}} with computational models to detect and interpret narrative features. The annotation scheme includes two components: (1) whether a text contains a story and (2) six narrative features that capture both the textual aspects of the story and its potential effects on readers. Using this dataset, we train and evaluate a series of encoder-based and generative classifiers. The best-performing models are then applied to explore how stories and their features appear in \textit{ChangeMyView} comments, focusing on differences between persuasive and non-persuasive arguments.

The main contribution of this work can be summarized as follows:
\begin{itemize}
    \item We introduce a framework for analyzing narrativity in online discussions, including a manually annotated \textit{ChangeMyView} dataset and supervised models for detecting narrativity and fine-grained narrative features, enabling large-scale narrativity analysis in \textit{ChangeMyView}.
        \item We compare encoder-based models with large language models in a zero-shot setting, evaluating performance for both scalar rating and binary detection of narrative features, and show that supervised models consistently outperform large language models on these tasks.
    \item We examine narrativity using both scalar (perspectivist) and binary (majority vote) representations, showing how modeling narrativity as a scalar property captures annotation variability and improves prediction compared to categorical approaches, also leading to more accurate results and conclusions when applied to downstream analyses.
    \item We analyze the role of fine-grained narrative features in online argumentation, providing insight into how different narrative dimensions contribute to persuasion in the \textit{ChangeMyView} community.
\end{itemize}

Data, code,\footnote{\url{https://github.com/saranabhani/ARGUS}} and models\footnote{https://huggingface.co/sara-nabhani/ARGUS-[ Story | Agency | Event-Sequncing | Surprise | Suspense | Curiosity ]} are publicly available.

%% file: ACM2026/sections/related_work.tex
\section{Related Work}

\subsection{Fine-grained Analysis of Narrative} Several studies have developed frameworks to computationally analyze fine-grained narrative features. \citet{Piper:2021} annotate text passages for \textit{Agency}, \textit{Event Sequencing}, and \textit{World Making}, informed by \citeauthor{herman2009basic}'s framework focusing on text-oriented features.
\citet{steg-etal-2022-computational}, informed by \citeauthor{sternberg:2003}'s theory focusing on perception-oriented features, annotate dimensions like \textit{Suspense}, \textit{Curiosity}, and \textit{Surprise} in texts from the same corpus. Both studies consider narrativity to be a multidimensional construct, and use these features accordingly as proxies of the local degree of narrativity a text has. Through narrativity-prediction experiments, they show that narrativity can be more precisely modeled as a scalar, local property of a text, rather than as a binary property. 
Based on these findings, in a later work \citet{piper-bagga-2025-narradetect} build a new dataset of passages annotated for narrativity %on a scalar scale 
and train classification models to detect it. They show that supervised models outperform large language models on their dataset, though with lower ability to generalize to other datasets.

\subsection{Narrative Impact on Argumentative Discussions}
\citet{falk-lapesa-2023-storyarg} introduce \textit{StoryARG}, a corpus annotated for both narrative features and argumentative elements. Their findings suggest that narratives proposing solutions are more persuasive, though the study's small corpus limits generalizability.
\citet{falk-lapesa-2024-stories} further apply the \textit{StoryARG} framework to Reddit data, identifying narrative types in various argumentative contexts, but without evaluating their persuasive impact.
\citet{antoniak-etal-2024-people} introduce \textit{StorySeeker} to detect narrative spans, finding that narratives are more prevalent in personal experience subreddits.
In a \textit{ChangeMyView} case study, they note that the presence of narrative does not strongly correlate with persuasive success. A similar conclusion was reached by \citet{nabhani-etal-2025-storytelling} in a large-scale analysis of \textit{ChangeMyView} data, examining narrative persuasion at multiple levels, including the comment, thread, and user persuasiveness skills.

Our work builds on an annotation framework similar to that of \citet{Piper:2021} and \citet{steg-etal-2022-computational}, and 
extends the studies of \citet{antoniak-etal-2024-people} and \citet{nabhani-etal-2025-storytelling} by examining how narratives and their features relate to persuasive outcomes in \textit{ChangeMyView}, providing a deeper understanding of the role of narrative in persuasion.

%% file: ACM2026/sections/data.tex
\section{Data and Annotation}

\subsection{Data}
\label{sec:data}
\begin{figure}[h]
  \centering
\includegraphics[width=0.8\linewidth]{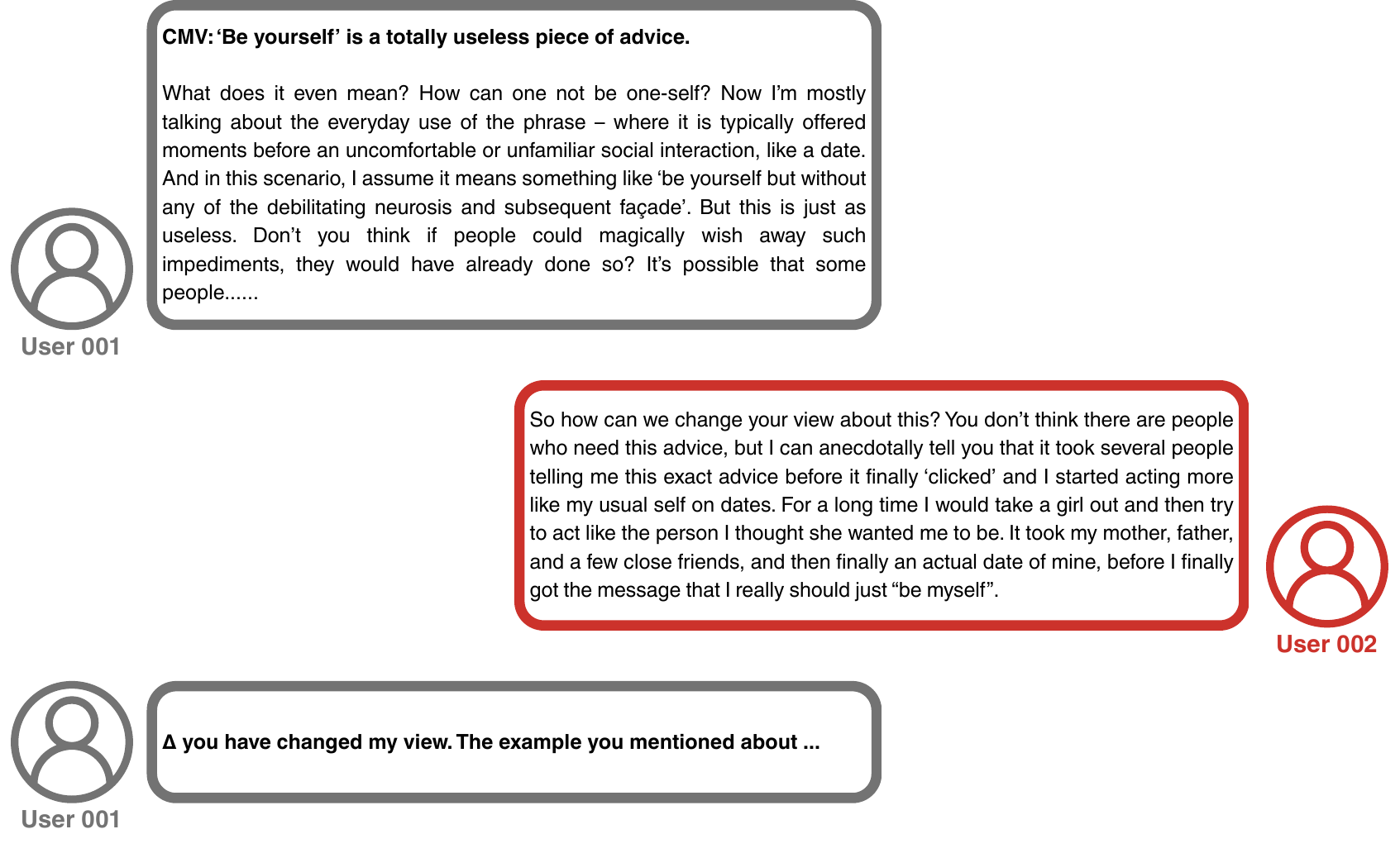}
    \caption{An example \textit{ChangeMyView} discussion thread. The original poster User 001 expresses a belief, and other users respond. User 002 received a delta, indicating their comment successfully changed the original poster's view.
    \label{fig:cmv_example}}
\end{figure}

To study the influence of narrative use in online argumentation, we focus on the \textit{ChangeMyView} forum. \textit{ChangeMyView} is an online community hosted on the Reddit platform where users discuss a wide range of topics by engaging with opposing viewpoints. In each discussion, a user, the original poster, presents a belief or opinion and invites others to provide arguments challenging their view. The original poster is expected to engage with responses within three hours resulting in an unstructured, multi-party debate.
If the original poster is persuaded by a comment, they reply to it with a \(\Delta\) symbol (delta), which serves as an explicit marker of persuasion and is usually accompanied by an explanation describing why their view has changed. Figure~\ref{fig:cmv_example} shows an example of a \textit{ChangeMyView} discussion. 

\textit{ChangeMyView} is well suited to our study because it is a public forum that closely resembles everyday, unstructured argumentation on social media platforms, and offers an explicit, user-generated signal of persuasion success through the \(\Delta\) mechanism.

We utilize an existing corpus of \textit{ChangeMyView} discussions constructed by \citet{tan+etal:16a}, which includes all \textit{ChangeMyView} threads posted between January 2013 and August 2015. Table~\ref{tab:dataset-stats} describes the corpus after excluding system- and moderator-generated threads.

\begin{table*}[h]
  \caption{Summary statistics of the \textit{ChangeMyView} dataset after preprocessing.}
  \label{tab:dataset-stats}
  \centering
  \begin{tabular}{lc}
    \toprule
    \textbf{Statistic} & \textbf{Value} \\
    \midrule
    Number of posts & 20,436 \\
    Number of unique original posters & 13,704 \\
    Number of comments & 1,017,724\\
    Number of awarded deltas & 11,643 \\
    \hline
    Average number of deltas per post & 0.57 \\
    Average number of comments per post & 49.96 \\
    Average number of unique participants per post & 23.74 \\
    \hline
    Average number of words per comment & 99.27\\
    Median number of words per comment & 59\\
    \bottomrule
  \end{tabular}
\end{table*}

To inform data selection for our study, we apply the story detection model from the \textit{StorySeeker} toolkit introduced by \citet{antoniak-etal-2024-people}.\footnote{\url{https://huggingface.co/mariaantoniak/storyseeker}}
 Based on the model's predictions, we select for annotation the 100 discussions with the highest proportion of comments classified as \textit{Story}, resulting in a total of 620 posts and comments. This sampling strategy enables us to focus on discussions with a clear presence of narrative content, rather than threads containing little or no storytelling.

\subsection{Narrative Annotation}

\subsubsection{Annotation Schema}
For story annotation, we use the codebook introduced in the \textit{StorySeeker} toolkit.\footnote{Original Story Codebook: \url{https://github.com/maria-antoniak/storyseeker}} However, our annotation setup differs in that we label each entire comment as either a \textit{Story} or a \textit{Non-Story}, whereas \textit{StorySeeker} also includes annotation of story and event spans. We therefore omit the span highlighting instructions and rely on the toolkit definitions of stories and the events that make them up.

\subsubsection{Agreement and Reliability}
Given the inherently subjective nature of story identification, different annotators may perceive the same comment differently, and multiple interpretations can be valid depending on perspective \cite{mire-etal-2024-empirical}. To capture multiple perspectives on what constitutes a story, we annotate the data with seven annotators. The annotators are undergraduate students with a solid background in Natural Language Processing (NLP) and an understanding of narrative theory. Before annotation, the annotators participated in a brief training session in which they discussed example texts and familiarized themselves with the annotation guidelines.

The overall inter-annotator agreement measured using Fleiss' $\kappa$ is 0.40. While this corresponds to fair agreement according to \citet{8d20e0b8-89d8-3d65-bcf5-8c19d56ec4ab}, such levels are expected for subjective tasks involving interpretation rather than objective labeling. However, to verify whether the fair level of agreement reflects systematic differences in perspective rather than annotation noise, we further analyze the annotations. We cluster annotators based on Cohen's $\kappa$ pairwise agreement, which results in two distinct groups. Within these groups, agreement increases to Fleiss' $\kappa$ values of 0.50 (moderate) and 0.61 (substantial), suggesting that annotators within each group share closer perspectives on story interpretation. 

To test the consistency of these perspectives, we rank annotators based on their strictness in story interpretation, measured by the proportion of items to which they assign a story label. We then randomly split the data into two subsets and compute rank correlations across the full set and both splits. This analysis yields a Spearman rank correlation of $\rho = 1$ with $p=0$, showing that the annotators' relative strictness is fully consistent across the dataset.
These results indicate that annotation disagreement reflects stable and meaningful differences in perspective rather than random error. Such perspectivist variation has been argued to be informative and useful for modeling subjective phenomena \cite{10.1609/aaai.v37i6.25840, plank-2022-problem}.

\subsection{Narrative Features Annotations}

\subsubsection{Annotation Schema}

\begin{figure*}[h]
  \centering
\includegraphics[width=0.8\linewidth]{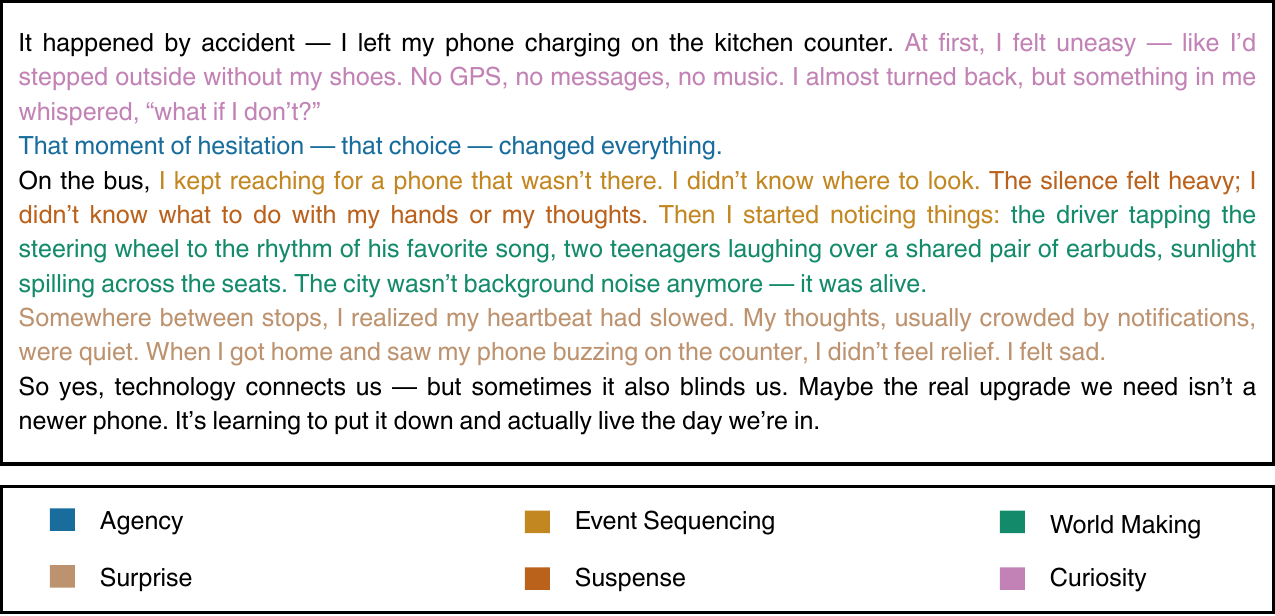}
    \caption{An example of \textit{Story} and Story features.
    \label{fig:story_example}}
\end{figure*}

For the annotation of narrative features, we rely on two complementary frameworks that study and operationalize narrativity through different sets of elements. The first framework is a narrativity model introduced by \citet{Piper:2021}, based on the theory by \citet{herman2009basic}, which defines the degree of narrativity using three textual elements: \textit{Agency}, \textit{Event Sequencing}, and \textit{World Making}.\footnote{Original Textual Elements Codebook: \url{https://doi.org/10.7910/DVN/DAWVME}} The second framework is the model introduced by \citet{steg-etal-2022-computational}, based on the three dimensions of reader perception proposed by \citet{sternberg:2003}, which captures narrativity from the reader's perspective through three perceptual elements: \textit{Suspense}, \textit{Curiosity}, and \textit{Surprise}.\footnote{Original Perceptual Elements Codebook: \url{https://github.com/KarloSlot/readers-perception-narrativity}}
Table~\ref{elements} lists all six elements with their descriptions, and Figure~\ref{fig:story_example} shows an example of their possible occurrence in argumentative texts.

\begin{table}[h]
\caption{Narrative features, their descriptions, and theoretical framework (Th.) -- \textbf{T}ext- or \textbf{R}eader oriented}
\label{elements}
\small
\begin{tabular}{p{0.3cm}p{2cm}p{11cm}}
\toprule
\textbf{Th.} &\textbf{Feature} & \textbf{Description} \\
\midrule
T & Agency & Extent to which a narrative centers on consistent, clearly defined characters driving the action. \\
%\hline
T & Event Sequencing & Temporal arrangement of events within the narrative. \\
%\hline
T & World Making & Construction of a fictional or realistic world through narrative elements. \\
\hline
R & Suspense & Presentation of information that suggests future events, thereby creating a delay in resolution. \\
%\hline
R & Curiosity & Presentation of information related to past events, leaving the reader intrigued by missing details. \\
%\hline
R & Surprise & Introduction of unexpected information about an event, eliciting a need for revising previous knowledge about the story. \\
\bottomrule
\end{tabular}
\end{table}

\subsubsection{Agreement and Reliability}
The same annotators who labeled texts as \textit{Story} or \textit{Non-Story} also carried out the annotation of the narrative features, but they were split into two groups. Three annotators labeled the textual elements; \textit{Agency}, \textit{Event Sequencing}, and \textit{World Making}. Whereas, given the higher expected subjectivity of the perceptual elements, \textit{Suspense}, \textit{Curiosity}, and \textit{Surprise}, were annotated by a larger group of four annotators.

Following the adopted frameworks by \citet{Piper:2021} and \citet{steg-etal-2022-computational}, which treat narrativity as a scalar property, we annotate all narrative elements using a five-point Likert scale ranging from 1 to 5, to capture the strength of the fine-grained narrative elements.

For inter-annotator reliability of the narrative features annotations, we report the Average Deviation Index (ADI) \cite{10.3389/fpsyg.2017.00777}, using both mean and median values, which is suitable for the five-point Likert scale data annotation. Given the subjective nature of the task, we also measure the consistency of the annotators using the Intraclass Correlation Coefficient, ICC(3,k). The results are shown in Table~\ref{tab:agreement_count}.

Agreement is strongest for the textual elements: \textit{Agency}, \textit{Event Sequencing}, and \textit{World Making} show low ADI values (mean ADI between 0.26 and 0.44), indicating that annotators typically differ by less than half a point on the Likert scale. These elements also achieve high ICC values reflecting strong consistency across annotators. The perceptual elements show slightly lower agreement. \textit{Suspense}, \textit{Curiosity}, and \textit{Surprise} have higher ADI values (mean ADI between 0.61 and 0.67), which is expected given their higher subjectivity. Similarly, ICC values for these elements are lower, but still indicate reasonable consistency. Overall, both ADI and ICC results support the reliability of the annotations.

\begin{table*}
\caption{Narrative Features Annotation Reliability and Consistency. Average Deviation Index (ADI; mean and median; range: [0–2]; lower is better) and Intraclass Correlation Coefficient (ICC(3,k); range: [0–1]; higher is better)}
\label{tab:agreement_count}
\centering
\begin{tabular}{l|cc}
\toprule
\textbf{Feature} & \textbf{ADI(mean/median)} $\downarrow$ & \textbf{ICC(3,k)} $\uparrow$ \\
\midrule
Suspense         & 0.612/0.555 & 0.720  \\
Curiosity        & 0.626/0.576 & 0.673  \\
Surprise         & 0.674/0.620 & 0.703 \\
\hline
Agency           & 0.428/0.338 & 0.852 \\
Event Sequencing & 0.440/0.356 & 0.831 \\
World Making     & 0.262/0.208 & 0.649  \\
\bottomrule
\end{tabular}
\end{table*}

\subsection{Corpus Description and Analysis}

\subsubsection{Association between Narrative Features and Story} 
We analyze how the six narrative features relate to the \textit{Story} label to understand whether these elements jointly characterize stories, maintaining internal coherence in the \textit{ChangeMyView} dataset. We apply linear and logistic regression models to examine this relationship.

We fit a linear regression to predict the \textit{Story} score (degree of narrativity) and a logistic regression to predict the presence of a \textit{Story}. The predictors are the six narrative feature scores, computed as the average of the annotators' ratings, controlling for the length of the text:
\begin{itemize}
    \item Linear Regression: \(Story_{scalar} \sim Agency_{scalar} + Event Sequencing_{scalar} + World Making_{scalar} + Surprise_{scalar} + Suspense_{scalar} + Curiosity_{scalar} + Text_{length}\)\\
    \item Logistic Regression: \(Story_{binary} \sim Agency_{scalar} + Event Sequencing_{scalar} + World Making_{scalar} + Surprise_{scalar} + Suspense_{scalar} + Curiosity_{scalar} + Text_{length}\)
\end{itemize}

Table~\ref{tab:analysis_scalar} shows that all six narrative features significantly and positively predict the overall \textit{Story} score. Similarly, Table~\ref{tab:analysis_binary} shows that narrative features predict the presence of a \textit{Story} in the argumenatative text, except for \textit{World Making} and \textit{Surprise}, which have no significant effect.

\begin{table}[h]
\centering
\caption{Linear regression predicting the \textit{Story} score from narrative features scores. $\boldsymbol{\beta}$= log-odds coefficients, \textbf{SE} = standard error, \textbf{t} = the t-test statistic, $\boldsymbol{p}$= p-value, and $\boldsymbol{\eta^2}$ = partial eta squared. The model's adjusted  $\boldsymbol{R^2}$= 0.74. Number of samples = 620.}
\label{tab:analysis_scalar}
\begin{tabular}{l|ccccc}
\toprule
\textbf{Predictor} & $\boldsymbol{\beta}$ & \textbf{SE} & \textbf{t}  &  $\boldsymbol{p}$ & $\boldsymbol{\eta^2}$ \\ 
\midrule
Agency\(_{scalar}\)      & 0.34 & 0.06 & 5.86 & $<0.001$ & 0.05 \\
Event Sequence\(_{scalar}\)      & 0.26 &  0.06 & 4.50 &$<0.001$ & 0.03 \\
World Making\(_{scalar}\)      & 0.07 & 0.03 & 2.80 &$0.005$ & 0.01  \\
Suspense\(_{scalar}\)      & 0.19 & 0.03  & 5.64 &$<0.001$ & 0.05 \\
Curiosity\(_{scalar}\)      & 0.17 & 0.03 & 5.39 &$<0.001$ & 0.05 \\
Surprise\(_{scalar}\)      & 0.10 &  0.03 & 2.92 &$0.004$ & 0.01\\
Text\(_{length}\)    & -0.08   & 0.03 & -2.90 &$0.004$ & 0.01\\
\bottomrule
\end{tabular}
\end{table}

\begin{table*}[h]
\centering
\caption{Logistic regression predicting the presence of \textit{Story} from narrative features scores. \textbf{$\beta$}= log-odds coefficients, \textbf{SE} = standard error, \textbf{z} = the Wald test statistic, $\boldsymbol{p}$= p-value, and \textbf{OR} = odds ratio (computed as $\exp(\beta)$). Number of samples = 620.}
\label{tab:analysis_binary}
\begin{tabular}{l|ccccc}
\toprule
\textbf{Predictor} & $\boldsymbol{\beta}$ & \textbf{SE}  & \textbf{z} & $\boldsymbol{p}$ & \textbf{OR}\\ 
\midrule
Agency\(_{scalar}\)      & 1.16 &  0.32 & 3.65 & $<0.001$ & 3.19\\
Event Sequencing\(_{scalar}\)      & 0.95 &  0.33 & 2.86 &   $0.004$ & 2.59\\
World Making\(_{scalar}\)      & 0.18 &  0.17 & 1.09 &  0.27 & 1.20 \\
Suspense\(_{scalar} \)     & 0.88 &  0.20 & 4.41 &  $<0.001$ & 2.41 \\
Curiosity\(_{scalar}\)      & 0.40 & 0.19  & 2.15 &  $0.03$ & 1.49\\
Surprise\(_{scalar}\)      & 0.33 & 0.20 & 1.68 &   0.09 & 1.39\\
Text\(_{length}\)    &  -0.77  & 0.172  &  -4.45 &   $<0.001$ &  0.46\\
\bottomrule
\end{tabular}
\end{table*}

\subsubsection{Narrativity Distribution}
We describe the dataset using both the presence and the strength of narrative features. Strength is derived from the variability in annotators' ratings and is computed as the average annotation value for each comment. This applies both to the Likert-scale ratings of narrative features and to the binary \textit{Story} labels.

\begin{figure*}[h]
  \centering
\includegraphics[width=0.8\linewidth]{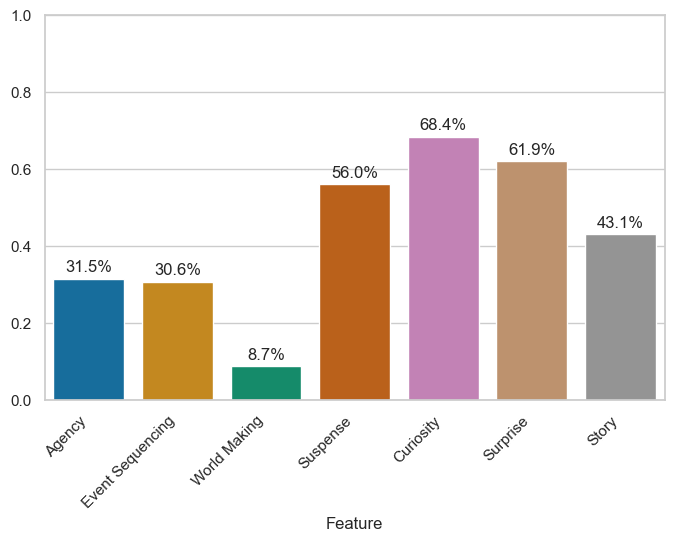}
    \caption{Distribution of the strength of Story occurrence and Story features.
    \label{fig:presence_dist}}
\end{figure*}

\begin{figure*}
    \centering
    \begin{subfigure}{\linewidth}
        \centering
        \includegraphics[width=0.8\linewidth]{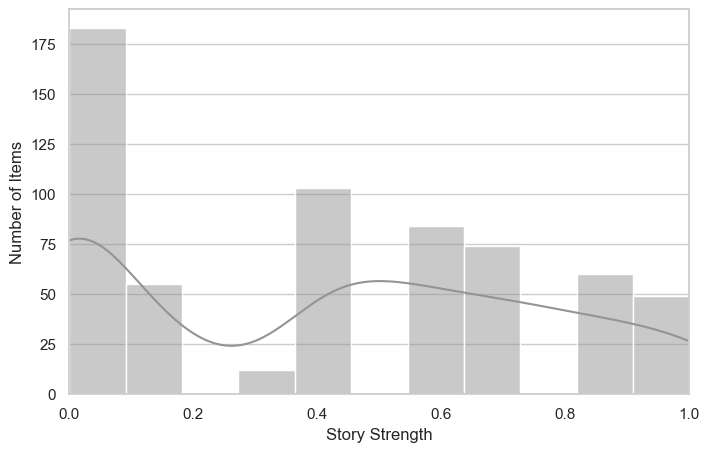}
        \caption{Story Occurrence Distribution}
    \end{subfigure}
    \vspace{1em}
    \begin{subfigure}{\linewidth}
        \centering
        \includegraphics[width=0.8\linewidth]{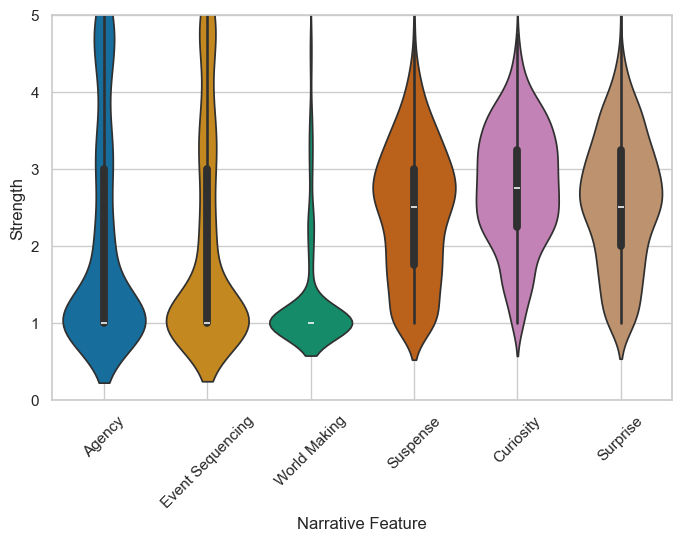}
        \caption{Narrative Features Distribution}
    \end{subfigure}

    \caption{Distribution of the features' strength scores calculated as the average of the annotators' ratings.}
    \label{fig:strength_dist}
\end{figure*}

To determine the presence of a story, we binarize these averaged values. For the narrative features, averaged ratings greater than or equal to 2.5 are marked as present, while for the \textit{Story} label, averaged values greater than or equal to 0.5 indicate the presence of a story. Overall, 43\% of the annotated comments are labeled as containing a story. The presence rates of \textit{Story} and the six narrative features are shown in Figure~\ref{fig:presence_dist}, and the distributions of strength values are provided in Figure~\ref{fig:strength_dist}.

Overall, the reader-oriented elements show stronger and higher presence than the textual elements. One possible explanation is the nature of discussions on \textit{ChangeMyView}, which are time-limited and goal-driven. In this setting, users may be more likely to rely on emotional influence than developing a structured narrative world.

\textit{World Making}, in particular, is relatively rare in our sampled data (8.7\%) and shows the weakest effects across the analysis. It does not significantly predict the presence of a story and has the smallest effect size on its score. Given its low prevalence and limited contribution to story identification in this dataset, we exclude it from the subsequent analysis.

%% file: ACM2026/sections/modeling.tex
\section{Modeling Narrativity}

We use the annotated corpus to train and evaluate models for detecting stories and narrative features, enabling large-scale analysis of narrative persuasion in \textit{ChangeMyView}.

\subsection{Data Splits and Label Representation}
For the \textit{Story} label and each narrative feature, the data is split into an 80\% subset (496 instances) used for training and validation and a 20\% subset (124 instances) held out for final evaluation. The splits are created using stratified sampling to ensure that label distributions are preserved across both subsets.

To represent the annotations, we use soft labels based on the distribution of annotators' ratings for each instance. For the \textit{Story} label, this distribution is computed over the annotators' binary ratings, while for the narrative features it is computed over the five-point Likert scale. This representation preserves variation across annotators and allows models to distinguish between clearly narrative instances and more ambiguous cases where perspectives differ.

By using soft labels, the model learns the strength of each narrative feature rather than only its presence, and captures plausible differences in interpretation instead of treating them as noise. This approach is consistent with previous works that model narrativity as a scalar property \citep{Piper:2021, steg-etal-2022-computational}, as well as with work showing the value of modeling annotation variability for subjective tasks where no single ground-truth label exists\cite{leonardelli-etal-2025-lewidi, 10.1609/aaai.v37i6.25840, plank-2022-problem, fornaciari-etal-2021-beyond}.

For the \textit{Story} label, we additionally train a second model variant using the binarized label. This allows us to compare performance with the soft-label variant and assess the benefit of explicitly modeling annotator variation.

\subsection{Model Selection and Evaluation}
Modeling of narrativity and the five fine-grained narrative features is carried out in two stages. In the first stage, we select the best model for each narrative feature among four encoder-based models: DistilBERT, BERT\(_{base} \), RoBERTa\(_{base} \), and ModernBERT\(_{base} \) via nested cross-validation. In the second stage, the best selected models in the first stage are then retrained and calibrated on the full training set. For models trained with soft labels, we evaluate predictions using the Brier score and Wasserstein distance to measure how closely the predicted probability distributions match the annotated rating distributions. We additionally report Root Mean Squared Error (RMSE) and Mean Absolute Error (MAE) to measure the difference between the predicted average score and the annotators' average rating, which reflects the error in the predicted strength of each narrative feature. For the binary \textit{Story} model trained with hard labels, we use Accuracy, F1, macro-F1, and weighted F1, which capture overall correctness as well as performance across class imbalance. Full model selection and validation details and results are in Appendix~\ref{sec:mod_select_eval}.

We evaluate the selected best-performing models on the 20\% held-out test set. For models trained with soft labels, we report both scalar and binary results. Scalar metrics measure how well models predict the strength of narrative features. For binary evaluation, we compute the expected Likert score from the predicted probability distribution by taking the weighted average of all Likert scale values, and then threshold this value to determine feature presence (0.5 for the \textit{Story} label and 2.5 for narrative features). For the \textit{Story} model trained with hard binary labels, we also report scalar metrics to compare its calibration with the soft-label variant. Test set results for the calibrated models are shown in Table~\ref{tab:test_results}.

\begin{table}[h]
\caption{Performance on the test set using the best fine-tuned model per feature (DistilBERT(D), BERT(B), RoBERTa(R), and ModernBERT(M)). Metric ranges are as follows: Brier score [0,2], Wasserstein distance and RMSE [0,1] for \textit{Story} and [0,4] for story features (lower is better);  Accuracy and F1 [0,1] (higher is better).}
\label{tab:test_results}
\centering
\begin{tabular}{l|c|ccc|cc}
\toprule
\textbf{Feature} & \textbf{Model} & \textbf{Brier} $\downarrow$ & \textbf{Wasserstein} $\downarrow$ & \textbf{RMSE} $\downarrow$ & \textbf{Accuracy} $\uparrow$  & \textbf{F1}$\uparrow$ \\ 
\midrule
Story\(_{binary}\)       & D  & 0.120 &  0.186 & 0.245 & 0.815 & 0.763 \\
Story\(_{scalar}\)       &  R & 0.076 & 0.150 & 0.194 & 0.879 & 0.842 \\
\hline
Suspense     &  B   & 0.188 & 0.573 & 0.554 & 0.774 & 0.771 \\
Curiosity    & R & 0.203  & 0.583 & 0.564 & 0.823 & 0.871\\
Surprise     & D & 0.207 & 0.635 & 0.612 & 0.758 & 0.795 \\
\hline
Agency       &   M    & 0.167 & 0.545 & 0.843 & 0.839 & 0.714 \\
Event Sequencing   & D & 0.146 & 0.554 & 0.771 & 0.871 & 0.765 \\
\bottomrule
\end{tabular}
\end{table}

Overall, the models show good performance in predicting narrativity both as a binary property and as a continuous score. Performance is highest for detecting the presence of \textit{Story} and \textit{Curiosity}. In general, reader-oriented features (\textit{Curiosity}, \textit{Suspense}, and \textit{Surprise}) have better performance, while text-oriented features (\textit{Agency} and \textit{Event Sequencing}) show better calibration.

For the \textit{Story} label, the model trained with soft labels consistently outperforms the binary model on both scalar and binary evaluation metrics. This finding aligns with prior work showing that subjective annotations do not have a single correct label and that models benefit from explicitly representing annotator variability \cite{10.1609/aimag.v36i1.2564, plank-2022-problem, 10.1609/aaai.v37i6.25840, fornaciari-etal-2021-beyond}. By modeling this variability, soft-label training improves calibration without sacrificing classification accuracy.

Results across narrative features further support this finding. Models trained with soft labels achieve strong performance on scalar metrics, indicating that they accurately capture the graded strength of narrative features. When these scalar predictions are thresholded to obtain binary decisions, performance remains high, demonstrating that the models preserve useful information about feature presence.

We also evaluate our \textit{Story} models against existing models from \textit{StorySeeker} \cite{antoniak-etal-2024-people} and \textit{NarrDetect} \cite{piper-bagga-2025-narradetect} to assess their generalizability and to compare performance with models trained on data from different domains. The results are reported in Appendix~\ref{sec:cross-datasets}. In addition, we compare the performance of our models on narrative detection and narrative features rating with large language models. The results show that fine-tuned encoder-based models outperform large language models on both tasks. Full results and prompting details are provided in Appendix~\ref{sec:llms_eval}.

%% file: ACM2026/sections/analysis.tex
\section{Narrativity and Persuasion}
\label{analysis}
\textit{Does narrative predict persuasion outcomes on ChangeMyView}? To test our main research question, we use the corpus of \citet{tan+etal:16a} (Section~\ref{sec:data}), taking \emph{Deltas} as proxies of persuasiveness. We exclude discussions and comments posted by moderators or system accounts as well as comments authored by users labeled as `[deleted]' in the corpus, obtaining 963,253 comments. 
On these, we apply our best-performing narrativity and narrative features models, For \textit{Story}, we use the scalar model given its higher performance and reliability.

The study is carried out in two steps:
\begin{enumerate}
    \item We test whether narrativity predicts persuasion.
    \item We examine which narrative features drive this effect.
\end{enumerate}
 
All analyses use mixed-effects logistic regression to estimate the probability that a comment receives a \emph{Delta}, including random intercepts for comment authors and original posters to account for repeated participation, and controlling for comment length. All continuous predictors are z-standardized.

Similar analyses could not be performed on manually-annotated \textsc{Argus} data because of its limited size, the rare occurrence of \emph{Delta} (2.3\%), and the very small repeated participation by the same users. This makes it difficult to reliably study the users' behaviour and the association between \emph{Delta} and narrativity use based on \textsc{Argus} data solely.    

\subsection{Are narrative arguments more likely to persuade?} First, we test if narrativity predicts persuasion success by modeling delta by narrativity score and comment length ($Model_1$),
and narrative presence (binarized using a cutoff of narrativity scores at a threshold of 0.5) and comment length ($Model_2$):
\begin{itemize}
    \item \(Model_{1}\): \(Delta \sim  Story_{scalar} + Text_{length} + (1|Author) + (1|OPAuthor)\) 
    \item \(Model_{2}\): \(Delta \sim  Story_{binary} + Text_{length} + (1|Author) + (1|OPAuthor)\)
\end{itemize}

Results, in Table~\ref{tab:analysis_m1_m2}, show that narrativity ($Model_1$) has a strong positive effect on \emph{Delta} ($\beta = 0.45$, $p < 0.001$): a one unit increase in narrativity raises persuasion odds by about 57\% (OR $\approx 1.57$). A model using the binary story label as predictor ($Model_2$) yields a comparable effect ($\beta = 0.62$, $p < 0.001$, OR $\approx 1.86$). Comment length also increases persuasion likelihood of \emph{Delta} ($\beta = 0.36$–$0.42$, OR $\approx 1.43$–$1.52$), but narrativity provides a robust persuasive advantage beyond comment length and differences between authors, which are controlled for in the model.

\begin{table}[h]
\caption{Results of \(Model_{1}\) and \(Model_{2}\). $\boldsymbol{\beta}$= log-odds coefficients, \textbf{SE} = standard error, \textbf{z} = the Wald test statistic, $\boldsymbol{p}$= p-value, and \textbf{OR} = odds ratio (computed as $\exp(\boldsymbol{\beta})$). Number of samples = 963,253.}
\label{tab:analysis_m1_m2}
\centering
\begin{tabular}{l|ccccc}
\toprule
\textbf{Predictor} & $\boldsymbol{\beta}$ & \textbf{SE}  & \textbf{z} & $\boldsymbol{p}$ & \textbf{OR} \\ 
\midrule
\multicolumn{6}{c}{\textbf{\(Story_{scalar} Predictor\) }}\\
\midrule
Intercept & -5.65 & 0.03 & 39.26 & $< 0.001$ &  - \\
Story\(_{scalar}\) & 0.45 & 0.01 & 39.26 & $< 0.001$ &  1.57 \\
Text\(_{length}\)    & 0.36 &  0.01 & 53.88 & $< 0.001$ & 1.52 \\
\midrule
\multicolumn{6}{c}{\textbf{\(Story_{binary} Predictor\) }}\\
\midrule
Intercept & -5.69 & 0.03 & -170.84 & $< 0.001$ &  - \\
Story\(_{binary}\) & 0.62 & 0.03 & 25.18 & $< 0.001$ & 1.86 \\
Text\(_{length}\)    & 0.42 & 0.01 & 67.31 & $< 0.001$ & 1.43\\
\bottomrule
\end{tabular}
\end{table}

\subsection{Which specific narrative features make a story persuasive?} 

Before testing what story features contribute to persuasion, we first verify that they reliably predict narrativity. From a theoretical perspective, each of the features in itself is not a necessary and sufficient property of narratives. Rather, it is their combined presence that increases the degree of narrativity~\cite{Pianzola2018Looking, herman2009basic}. Accordingly, we create two composite measures: a \textit{Structural} score, defined as the average of \textit{Agency} and \textit{Event Sequencing}, and a \textit{Response} score, defined as the average of \textit{Suspense}, \textit{Curiosity}, and \textit{Surprise}. Using these predictors and controlling for comment length, we estimate two models, one predicting a story presence ($Model_4$)and one predicting a continuous narrativity score ($Model_3$):

\begin{itemize}
    \item \(Model_{3}\): \(Story_{scalar} \sim Structural_{score} + Response_{score} + Text_{length}\)
    \item \(Model_{4}\): \(Story_{binary} \sim Structural_{score} + Response_{score} + Text_{length}\)
\end{itemize}

Results are reported in Table~\ref{tab:analysis_m3_m4}. Both \textit{Structural} and \textit{Response} scores are strong positive predictors of narrativity. In the logistic model predicting the binary story label, increases in both composite scores substantially raise the likelihood that a text is classified as a story, while text length is negatively associated with story classification. In the linear model predicting the continuous narrativity score ($Model_3$), \textit{Structural} and \textit{Response} scores again show large positive effects, confirming that both composite narrativity scores capture the core dimensions of narrativity, but they do not completely overlap~\cite{steg-etal-2022-computational}. 

\begin{table}[h]
\caption{Results of \(Model_{3}\) and \(Model_{4}\). $\boldsymbol{\beta}$= log-odds coefficients, \textbf{SE} = standard error, \textbf{z} = the Wald test statistic, \textbf{t} = the t-test statistic, $\boldsymbol{p}$= p-value, \textbf{OR} = odds ratio (computed as $\exp(\boldsymbol{\beta})$), and $\boldsymbol{\eta^2}$ = partial eta squared. \(Model_{3}\) adjusted $\boldsymbol{R^2}$= 0.77. Number of samples = 963,253.}
\label{tab:analysis_m3_m4}
\centering
\begin{tabular}{l|ccccc}
\toprule
\multicolumn{6}{c}{\textbf{ \(Story_{scalar} \) }}\\
\midrule
\textbf{Predictor} & $\boldsymbol{\beta}$ & \textbf{SE}  & \textbf{t} & $\boldsymbol{p}$ & $\boldsymbol{\eta^2}$ \\ 
\midrule
Intercept & 0.00 & 0.00 & 0.00 &1.00  &  - \\
Response\(_{score}\) & 0.49 &0.00 &715.34 & $< 0.001$ & 0.35  \\
Structural\(_{score}\) & 0.53 & 0.00&766.55&$< 0.001$ &  0.38 \\
Text\(_{length}\)    & -0.06 &  0.00 &-80.05&$< 0.001$  & 0.01\\
\hline
\multicolumn{6}{c}{\textbf{\(Story_{binary} \)}}\\
\midrule
\textbf{Predictor} & $\boldsymbol{\beta}$ & \textbf{SE}  & \textbf{z} & $\boldsymbol{p}$ & \textbf{OR} \\ 
\hline
Intercept & -2.76 & 0.01 & -393.28 & $< 0.001$ & -  \\
Response\(_{score}\) & 1.17 & 0.01 & 130.49 & $< 0.001$ & 3.22 \\
Structural\(_{score}\) & 1.91 &  0.01& 275.40 & $< 0.001$& 6.75 \\
Text\(_{length}\)    & -0.39 & 0.00 & -84.55 & $< 0.001$ & 0.68\\
\bottomrule
\end{tabular}
\end{table}

Having established this, we next examine how these composite scores contribute to persuasive outcomes, using the following models:

\begin{itemize}
    \item \(Model_{5}\): \(Delta \sim Structural_{score} + Response_{score} + Text_{length} + (1|Author) + (1|OPAuthor)\)
    \item \(Model_{6}\): \(Delta \sim Structural_{binary} + Response_{binary} + Text_{length} + (1|Author) + (1|OPAuthor)\)
\end{itemize}

Results are reported in Table~\ref{tab:analysis_m5_m6}. Both narrative dimensions are positively associated with persuasion. The \textit{Structural} score has a small but significant positive effect ($\beta = 0.09$, $p < 0.001$), while the \textit{Response} score shows a much larger effect ($\beta = 0.59$, $p < 0.001$). Text length remains a strong predictor of persuasion ($\beta = 0.30$, $p < 0.001$). In terms of effect sizes, a one-unit increase in response score increases the odds of persuasion by about 80\% (OR $\approx 1.80$), while a one-unit increase in structural score increases the odds by about 9\% (OR $\approx 1.09$). 

% TEMP

\begin{table}[h]
\caption{Results of \(Model_{5}\) and \(Model_{6}\). $\boldsymbol{\beta}$= log-odds coefficients, \textbf{SE} = standard error, \textbf{z} = the Wald test statistic, $\boldsymbol{p}$= p-value, and \textbf{OR} = odds ratio (computed as $\exp(\boldsymbol{\beta})$). Number of samples = 963,253.}
\label{tab:analysis_m5_m6}
\centering
\begin{tabular}{l|ccccc}
\toprule
\textbf{Predictor} & $\boldsymbol{\beta}$ & \textbf{SE}  & \textbf{z} & $\boldsymbol{p}$ & \textbf{OR} \\ 
\midrule
\multicolumn{6}{c}{Scalar Predictors}\\
\midrule
Intercept & -5.70 & 0.03 & -175.32 & $< 0.001$ &  - \\
Response\(_{score}\) & 0.59 & 0.02 & 30.24 & $< 0.001$ & 1.80 \\
Structural\(_{score}\) & 0.09 & 0.01 & 7.80 & $< 0.001$ & 1.09\\
Text\(_{length}\) & 0.30 & 0.01 & 39.91 & $< 0.001$ &  1.35\\
\midrule
\multicolumn{6}{c}{Binary Predictors}\\
\midrule
Intercept & -6.20 & 0.04 & -165.33 & $< 0.001$ &  - \\
Response\(_{binary}\) & 1.09 & 0.03 & 37.18 & $< 0.001$ & 2.97 \\
Structural\(_{binary}\) & 0.34 & 0.03 & 11.59 & $< 0.001$ & 1.40\\
Text\(_{length}\) & 0.35 & 0.01 & 51.74 & $< 0.001$ & 1.42 \\
\bottomrule
\end{tabular}
\end{table}

We then examine the effects of individual narrative features on persuasion by predicting \emph{Delta} using models that include each fine-grained feature separately:

\begin{itemize}
    \item \(Model_{7}\): \(Delta \sim Agency_{scalar} + Event Sequencing_{scalar} + Surprise_{scalar} + Suspense_{scalar} + Curiosity_{scalar} + Text_{length} + (1|Author) + (1|OPAuthor)\)
 \item \(Model_{8}\): \(Delta \sim Agency_{binary} + Event Sequencing_{binary} + Surprise_{binary} + Suspense_{binary} + Curiosity_{binary} + Text_{length} + (1|Author) + (1|OPAuthor)\)
\end{itemize}

Results are reported in Table~\ref{tab:analysis_m7_m8}. When entered as scalar predictors ($Model_7$), all features show a significant association with persuasion. \textit{Curiosity} has the strongest effect ($\beta = 0.60$, $p < 0.001$, OR $\approx 1.82$), followed by \textit{Suspense} ($\beta = 0.15$, $p < 0.001$, OR $\approx 1.16$). \textit{Event Sequencing} ($\beta = 0.06$, $p < 0.001$) and \textit{Agency} ($\beta = 0.03$, $p = 0.01$) also show a positive association with persuasion, however, with smaller effect size (\textit{Event Sequencing}: OR $\approx 1.06$, \textit{Agency}: OR $\approx 1.03$). Whereas \textit{Surprise} exhibits a significant negative effect ($\beta = -0.11$, $p = 0.005$).

When narrative features are instead modeled as binary indicators ($Model_8$), the pattern changes. In this case, the presence of \textit{Surprise} shows a large positive association with persuasion (OR $\approx 1.51$), while the effects of \textit{Event Sequencing}, \textit{Agency}, and \textit{Suspense} are more modest. \textit{Curiosity} again shows the largest effect (OR $\approx 2.05$).

\begin{table}[h]
\caption{Results of \(Model_{7}\) and \(Model_{8}\). $\boldsymbol{\beta}$= log-odds coefficients, \textbf{SE} = standard error, \textbf{z} = the Wald test statistic, $\boldsymbol{p}$= p-value, and \textbf{OR} = odds ratio (computed as $\exp(\boldsymbol{\beta})$). Number of samples = 963,253.}
\label{tab:analysis_m7_m8}
\centering
\begin{tabular}{l|ccccc}
\toprule
\textbf{Predictor} & $\boldsymbol{\beta}$ & \textbf{SE}  & \textbf{z} & $\boldsymbol{p}$ & \textbf{OR} \\ 
\midrule
\multicolumn{6}{c}{Scalar Predictors}\\
\midrule
Intercept & -5.72 & 0.03 & -175.572 & $< 0.001$ &  -\\
Agency\(_{scalar}\)      & 0.03 & 0.01  & 2.453 & 0.01 & 1.03 \\
Event Sequencing\(_{scalar}\)      & 0.06 & 0.02  & 3.59 & $< 0.001$ & 1.06 \\
Suspense\(_{scalar}\)      & 0.15 & 0.03  &4.90&$< 0.001$ & 1.16\\
Curiosity\(_{scalar}\)      & 0.60 & 0.04 & 15.865 & $< 0.001$ & 1.82 \\
Surprise\(_{scalar}\)      & -0.11 & 0.04 & -2.83 & $0.005$ &  0.89 \\
Text\(_{length}\)    &  0.30 & 0.01 & 37.56 & $< 0.001$ & 1.35 \\
\midrule
\multicolumn{6}{c}{Binary Predictors}\\
\midrule
Intercept & -6.43 & 0.04 & -151.452 & $< 0.001$ & -\\
Agency\(_{binary}\)  & 0.16 & 0.04 & 4.37 & $< 0.001$ & 1.17 \\
Event Sequencing\(_{binary}\)      & 0.17 & 0.03  & 5.07 & $< 0.001$ & 1.19 \\
Suspense\(_{binary}\)      & 0.30 & 0.03  & 9.85 & $< 0.001$ & 1.35  \\
Curiosity\(_{binary}\)      & 0.72 & 0.04 & 16.28 & $< 0.001$ & 2.05 \\
Surprise\(_{binary}\)      & 0.41 &  0.04 &10.55& $< 0.001$ & 1.51 \\
Text\(_{length}\)    & 0.33 & 0.01 & 45.49 & $< 0.001$ & 1.39  \\
\bottomrule
\end{tabular}
\end{table}

\subsection{Discussion}

\subsubsection{Narrativity as a scalar rather than a binary property}

\begin{figure*}
    \centering
    \begin{subfigure}{\linewidth}
        \centering
        \includegraphics[width=0.8\linewidth]{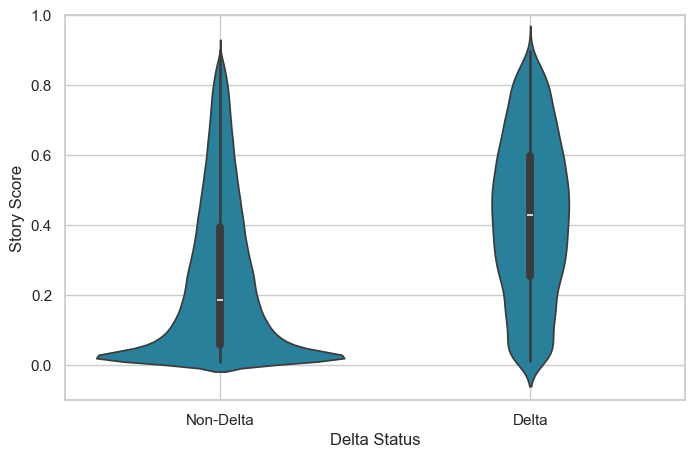}
        \caption{The distribution of story score per delta status}
        \label{fig:story_rs_dist_story}
    \end{subfigure}
    \vspace{1em}
    \begin{subfigure}{\linewidth}
        \centering
        \includegraphics[width=0.8\linewidth]{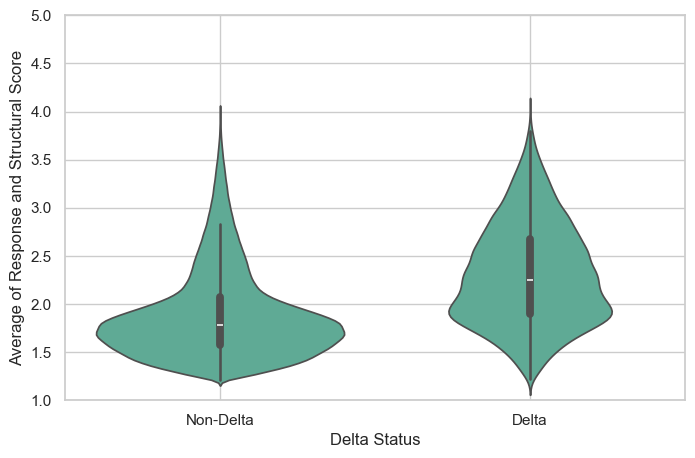}
        \caption{The distribution of the narrativite composites average per delta score}
        \label{fig:story_rs_dist_rs}
    \end{subfigure}

    \caption{The distribution of narrativity scores per delta status in the \textit{ChangeMyView} corpus.}
    \label{fig:story_rs_dist}
\end{figure*}

Although the analyses show that both scalar and binary representations of narrativity predict persuasion, the distribution (Figure~\ref{fig:story_rs_dist}) and the regression results indicate that the binary representation loses information. Persuasive comments (Delta) have an approximately normal distribution of narrativity scores. Whereas the non-persuasive comments (Non-Delta) are heavily concentrated at very low narrativity levels, with the median value $<0.2$ and the density decreasing as narrativity increases. This shows that there is no clear narrativity cutoff point that separates persuasive from non-persuasive comments. The same is shown by Figure~\ref{fig:story_rs_dist_rs}, where the non-persuasive comments cluster at lower narrativity sub-dimensions scores ($1.2-2$), while the persuasive comments' density is highest around moderate scores ($2-3$), with a large overlap with the distribution of the non-persuasive comments. Both distributions, in addition to the regression results, suggest that persuasion is associated with scalar increases in narrativity rather than a binary transition from \textit{Non-Story} to \textit{Story}. Using a fixed threshold---such as the probability cutoff of 0.5 used in many NLP approaches that model narrativity as a binary property---would treat comments that are just below and just above the threshold as different categories (such as comments with 0.49 and 0.51 narrativity score), while group together comments with a higher difference in narrativity (such as comments with 0.51 and 0.90 narrativity score) losing information that contributes to the explanation of the association between narrativity and persuasion, and therefore leading to a less accurate interpretation. This loss of information can also be seen in the analyses results (Table~\ref{tab:analysis_m1_m2},~\ref{tab:analysis_m5_m6}, and~\ref{tab:analysis_m7_m8}), when comparing the effect size of the narrativity sub-dimensions and individual features when used as binary predictors, in contrast to when used as scalar predictors of persuasion. When modeled as binary predictors, narrative features consistently show larger odds ratios than when used as scalar predictors. 
% \textit{Surprise} shows a contrasting pattern. When modeled as a scalar predictor, it has a small negative effect on persuasion, while when modeled as a binary predictor, it shows a large positive effect, suggesting that some information on the association with persuasion were not captured due to the binary grouping of the comments' surprise scores. 
For example, when modeled as a scalar predictor, \textit{Surprise} has a small negative effect on persuasion, while when modeled as a binary predictor, it shows that the presence of \textit{Surprise} has a large positive effect on persuasion, not capturing that the effect decreases as the degree of \textit{Surprise} increases due to the binary grouping of the comments' \textit{Surprise} scores. 

\subsubsection{Narrativity as a scalar predictor of persuasion}
Across all analyses, narrativity is consistently associated with persuasion on \textit{ChangeMyView}. Comments that are higher with narrativity are more likely to receive a \emph{Delta}, even after controlling for comment length and differences between authors and original posters. This pattern holds whether narrativity is modeled as an overall story score, through composite measures, or through individual narrative features. The comment length is a strong positive predictor across all of the analyses.  This finding differs from prior conclusions which find weaker or more limited effects of narrativity on persuasion: \citet{antoniak-etal-2024-people}---who focus only on the binary presence of ``a sequence of events involving one or more people''---and \citet{nabhani-etal-2025-storytelling}---who used the same classifier but also looked at predictions' confidence scores. 

Examining narrativity at the feature level helps identify which aspects of stories are most effective. When using composite measures, both the \textit{Structural} dimensions (\textit{Agency} and \textit{Event Sequencing}) and the \textit{Response} dimensions (\textit{Suspense}, \textit{Curiosity}, and \textit{Surprise}) are positively related to persuasion. The \textit{Response} dimension shows a much stronger effect, with a one-unit increase in \textit{Response} score increasing the odds of persuasion by 80\%. The \textit{Structural} dimension also contributes, but its effect is very modest compared to the effects of the \textit{Response} dimension and comment length.

The fine-grained analysis shows clear differences between individual narrative features. The features related to reader-response have larger effects than the structural features on persuasion, with \textit{Curiosity} being the largest contributor, followed by \textit{Suspense}. \textit{Surprise}, however, has a significant negative effect. Both structural features have a limited effect, with the contribution of \textit{Event Sequencing} slightly higher and stronger than that of \textit{Agency}. These findings suggest that persuasive success is driven more by how a comment engages the reader by managing how information is disclosed, rather than by how its narrative is structured. Using \textit{Curiosity} and \textit{Suspense} to engage readers by encouraging them to seek missing information and anticipate how the story will unfold are particularly effective for persuasion, whereas introducing unexpected elements may disrupt the engagement and reduce the persuasive effect.   

Overall, the findings show that narrativity is a meaningful predictor of persuasion, with structure and reader engagement contributing in different ways.

%% file: ACM2026/sections/conclusion.tex
\section{Conclusion}

This paper examines narrativity and its role in persuasion in the \textit{ChangeMyView} community. We introduce \textsc{Argus}, a framework that combines an annotated dataset of \textit{ChangeMyView} discussions, with supervised models to detect narrativity and five fine-grained narrative features: \textit{Agency}, \textit{Event Sequencing}, \textit{Suspense}, \textit{Curiosity}, and \textit{Surprise}. For the supervised models, we test multiple encoder-based models: DistilBERT, BERT\(_{base} \), RoBERTa\(_{base} \), and ModernBERT\(_{base} \). However, none of the models shows a significantly higher performance than the others.

Overall, \textsc{Argus} models show good performance predicting and rating narrative and narrative features, both the text- and reader-oriented features, with a stronger performance on the reader-oriented features---such as \textit{Suspense}, \textit{Curiosity}, and \textit{Surprise}---than the text-oriented features that build the structure of the story.

We also compare the performance of \textsc{Argus} fine-tuned encoder-based models with LLMs (Llama 3.1 8B and 70B) in a zero-shot setting. Across all narrative features, the fine-tuned encoder-based models consistently outperform Llama models in both binary detection and scalar rating of narrativity and narrative features, even when using the larger variant of Llama.

Given the subjectivity of narrativity perception, we also examine the benefits of modeling narrativity as a scalar rather than a binary property, both in terms of prediction performance and downstream analyses. We compare both soft-label (scalar) and hard-label (binary) representations of narrative annotations, showing that exposing the model to the annotation variability leads to more reliable predictions than providing a single majority-vote label. This is measured by metrics assessing both calibration and prediction performance of the models when operating as binary classifiers or as scalar raters. The analysis of narrative persuasion in \textit{ChangeMyView} showed that when treating narrativity and narrative features as binary predictors of persuasion, the results do not capture detailed information on their association. Specifically, the sharp score cutoff imposed by binarizing narrativity discards the persuasive effects of the comments with narrativity scores below the threshold, categorizing them as \textit{Non-Story}. Whereas persuasion can span over a wide range of continuous narrativity scores, with its success appears to rather benefit from scalar increases in narrativity than from belonging to discrete \textit{Story} and \textit{Non-Story} groups.

To study the impact of narrativity on persuasion in \textit{ChangeMyView}, we apply \textsc{Argus} scalar-based models to large-scale data from \textit{ChangeMyView}, showing that narrativity is an effective predictor of persuasion. Increases in the degree of narrativity are associated with higher persuasive success, in contrast to findings reported in previous studies. Narrative features contribute in distinct ways. Both features related to the structure of the story and those related to the reader response are associated with persuasion, with the response ones having a larger and stronger effect. Among individual narrative features, \textit{Curiosity} and \textit{Suspense} show the strongest effects on persuasive success. However, higher degrees of \textit{Surprise} are associated with reduced persuasion. While \textit{Event Sequencing} and \textit{Agency} show a very modest effect on persuasion compared to the features related to the readers' response.

% \subsection{Limitations and Future Works}
Our comparison with LLMs is limited to models from the Llama family and to zero-shot prompting. These results may not generalize to other model families or to different prompting strategies. Future work could evaluate models from other families and examine whether few-shot prompting or fine-tuning can improve LLMs' performance in detecting narrativity.

Our analyses focus on exploring the association between narrative and persuasion, and which dimensions of narrativity contribute more to the persuasion success. Further analyses could look into the interaction between the two narrativity dimensions (response and structural), in addition to the interaction of reader and text-oriented features, and their effect on persuasion outcome. 

Future work could also explore how narrativity varies across discussion topics and whether specific narrative features are employed differently. For example, topics related to values or social issues may rely more on agency and personal stories than technical or scientific topics and might make more use of emotionally engaging features like \textit{Surprise}, \textit{Suspense}, or \textit{Curiosity}. Further analysis could also study whether successful persuaders in \textit{ChangeMyView}, measured by the number of \emph{Deltas} awarded, show a systematic pattern in their use of narrativity in comparison to other users, and whether narrativity use increases as users gain more experience contributing to the forum.

%% file: ACM2026/sections/appendix.tex
\section{Modeling Results and Discussion}
\label{sec:mod_res_disc}

\subsection{Model Selection and Evaluation}
\label{sec:mod_select_eval}
Modeling of narrativity and the five fine-grained narrative features is carried out in two stages.
\paragraph{Stage 1: Model selection via nested cross-validation.}
We first perform nested cross-validation with five outer folds and three inner folds. The inner folds are used for hyperparameter tuning, while the outer folds are used to estimate generalization performance. For each feature, we evaluate four encoder-based models: DistilBERT, BERT\(_{base} \), RoBERTa\(_{base} \), and ModernBERT\(_{base} \). Performance is averaged across the outer folds.

For models trained with soft labels, we evaluate predictions using the Brier score and Wasserstein distance to measure how closely the predicted probability distributions match the annotated rating distributions. We additionally report Root Mean Squared Error (RMSE) and Mean Absolute Error (MAE) to measure the difference between the predicted average score and the annotators' average rating, which reflects the error in the predicted strength of each narrative feature. For the binary \textit{Story} model trained with hard labels, we use Accuracy, F1, macro-F1, and weighted F1, which capture overall correctness as well as performance across class imbalance.

Statistical testing on fold-level results, using the Friedman test followed by pairwise Wilcoxon tests, shows no significant differences between the models. We therefore select, for each feature, the model that achieves the best overall rank across evaluation metrics, taking into account both performance and stability across folds. Full validation results are reported in Table~\ref{tab:cv_results_soft} for scalar models and in Table~\ref{tab:cv_results_hard} for the binary \textit{Story} model.

\paragraph{Stage 2: Final training and calibration.}
In the second stage, the selected model for each feature is retrained on 80\% of the full training split, using the mean or mode of the optimal hyperparameters identified during cross-validation. The remaining 20\% of the training data is used for temperature scaling to calibrate the predicted probability distributions. Calibration is important because the subsequent analysis relies on the models' predictions to represent the strength and presence of narrative features.

\begin{table*}[h]
\caption{Performance comparison of soft-label trained DistilBERT(D), BERT(B), RoBERTa(R), and ModernBERT(M) models across 5-fold cross-validation. Results are reported as mean and standard deviation for each evaluation metric. Metric ranges are as follows: Brier score [0,2], Wasserstein distance,RMSE, and MAE [0,1] for \textit{Story} and [0,4] for story features. The best-performing model for each feature is highlighted in bold (lower is better).}
\label{tab:cv_results_soft}
% \small
\centering
\begin{tabular}{l|c|cc|cc}
\toprule
\textbf{Feature} & \textbf{Model} & \textbf{Brier\(_{AVG/STD}\)}$\downarrow$   & \textbf{Wasserstein\(_{AVG/STD}\)}$\downarrow$ & \textbf{RMSE\(_{AVG/STD}\)}$\downarrow$ & \textbf{MAE\(_{AVG/STD}\)}$\downarrow$ \\ 
\midrule
\multirow{4}{*}{Story\(_{scalar}\)}  & D  & 0.084/0.011 & 0.161/0.010 & 0.204/0.0136 & 0.161/0.010\\
       & B  & 0.103/0.017 & 0.179/0.017 & 0.226/0.020 & 0.179/0.017 \\
       & \textbf{R}  & 0.100/0.008 & 0.173/0.013 & 0.223/0.009 & 0.173/0.013 \\
       & M  & 0.091/0.010 & 0.161/0.010 & 0.213/0.012 & 0.161/0.010 \\
\hline
\multirow{4}{*}{Suspense}  & D  & 0.196/0.020 & 0.639/0.059 & 0.607/0.055 & 0.493/0.052 \\
       & \textbf{B}  & 0.196/0.014 & 0.629/0.043 & 0.610/0.056 & 0.491/0.049 \\
       & R  & 0.195/0.016 & 0.609/0.056 & 0.595/0.056 & 0.469/0.041 \\
       & M  & 0.200/0.012 & 0.611/0.056 & 0.603/0.055 & 0.464/0.049 \\
\hline
\multirow{4}{*}{Curiosity}  & D  & 0.211/0.017 & 0.651/0.023 & 0.574/0.019 & 0.467/0.017 \\
       & B  & 0.218/0.016 & 0.672/0.024 & 0.609/0.032 & 0.492/0.035 \\
       & \textbf{R}  & 0.224/0.009 & 0.659/0.019 & 0.608/0.021 & 0.489/0.015 \\
       & M  & 0.223/0.014 & 0.650/0.033  & 0.583/0.024 & 0.463/0.031 \\
\hline
\multirow{4}{*}{Surprise}  & \textbf{D}  & 0.206/0.019  & 0.699/0.044 & 0.653/0.063  & 0.537/0.047 \\
       & B  & 0.210/0.018 & 0.704/0.046 & 0.664/0.079 & 0.547/0.052 \\
       & R  & 0.203/0.012 & 0.670/0.035 & 0.637/0.065 & 0.503/0.053 \\
       & M  & 0.207/0.021 & 0.662/0.052 & 0.642/0.069 & 0.516/0.052 \\
\hline
\multirow{4}{*}{Agency}  & D  & 0.179/0.023 & 0.655/0.102 & 0.838/0.093 & 0.601/0.099 \\
       & B  & 0.213/0.053 & 0.823/0.161 & 0.961/0.140 & 0.780/0.159\\
       & R  & 0.207/0.035 & 0.739/0.144  & 0.960/0.127 & 0.689/0.145 \\
       & \textbf{M}  & 0.183/0.020 & 0.617/0.045 & 0.845/0.094 & 0.578/0.057 \\
\hline
\multirow{4}{*}{Event Sequencing}  & \textbf{D}  & 0.184/0.024 & 0.748/0.086 & 0.930/0.094 & 0.708/0.088 \\
       & B  & 0.201/0.043  & 0.792/0.140 & 0.968/0.143 & 0.739/0.139 \\
       & R  & 0.190/0.030 & 0.730/0.095 & 0.967/0.118 & 0.689/0.098 \\
       & M  & 0.199/0.033  & 0.739/0.084 & 0.946/0.109 & 0.701/0.089 \\
\bottomrule
\end{tabular}
\end{table*}

\begin{table*}[h]
\caption{Performance comparison of hard-label \textit{Story} trained BERT(B), DistilBERT(D), RoBERTa(R), and ModernBERT(M) models across 5-fold cross-validation. Results are reported as mean and standard deviation for each evaluation metric. The best-performing model is highlighted in bold (higher is better).}
\label{tab:cv_results_hard}
\centering
\begin{tabular}{l|c|cccc}
\toprule
\textbf{Feature} & \textbf{Model} & \textbf{Accuracy\(_{AVG/STD}\)}$\uparrow$  & \textbf{F1\(_{AVG/STD}\)}$\uparrow$  &  
\textbf{macro-F1\(_{AVG/STD}\)}$\uparrow$  & \textbf{weightd-F1\(_{AVG/STD}\)}$\uparrow$ \\ 
\midrule
\multirow{4}{*}{Story\(_{binary}\)}  & \textbf{D}  & 0.800/0.034 & 0.765/0.033 & 0.795/0.033 &  0.799/0.033 \\
& B  & 0.796/0.041 &0.769/0.045 & 0.793/0.041 & 0.797/0.041 \\
       & R  & 0.799/0.060 & 0.763/0.076 & 0.794/0.063 & 0.798/0.061 \\
       & M  & 0.768/0.050 & 0.725/0.053 & 0.762/0.049 & 0.767/0.049 \\
\bottomrule
\end{tabular}
\end{table*}

As shown in the cross-validation results, differences between model architectures are small and not statistically significant. DistilBERT, BERT, RoBERTa, and ModernBERT achieve comparable performance, and the task does not appear to benefit from ModernBERT's larger context window.

\subsection{Cross-Datasets Generalization}
\label{sec:cross-datasets}
We test our fine-tuned story detection models, both scalar and binary, on the \textit{StorySeeker} \cite{antoniak-etal-2024-people} and \textit{NarrDetect} \cite{piper-bagga-2025-narradetect} test datasets. We also evaluate the models released with these datasets on our annotated test set. This test allows us to examine how well models generalize across different data domains. 

\begin{figure}[h]
  \centering
\includegraphics[width=0.5\linewidth]{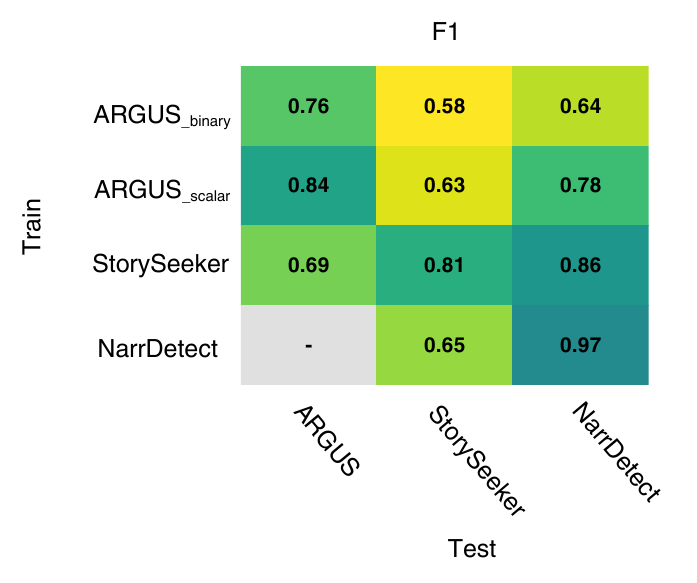}
    \caption{Cross-dataset performance of story detection models. Models fine-tuned on our data are evaluated on the \textit{StorySeeker} and \textit{NarrDetect} test sets, and the model released with the \textit{StorySeeker} toolkit is evaluated on our annotated test set. The \textit{NarrDetect} model is not publicly available and therefore could not be evaluated on our dataset.
    \label{fig:cross-data-perf}}
\end{figure}

Results, shown in Figure~\ref{fig:cross-data-perf}, indicate that narrativity models generalize reasonably well across datasets but are sensitive to domain differences. Models fine-tuned on our data perform well on the \textit{NarrDetect} test set (F1 = 0.78), although performance is lower than in-domain results. Performance on the \textit{StorySeeker} test set is comparable to that of \textit{NarrDetect}. In contrast, the model released with the \textit{StorySeeker} dataset shows reduced performance when evaluated on our annotated test set. The \textit{NarrDetect} model is not publicly available and therefore could not be evaluated on our dataset. Overall, these results highlight the importance of in-domain training for good-performing narrativity detection.

\subsection{Evaluation of LLMs}
\label{sec:llms_eval}
We additionally evaluate LLMs in a zero-shot setting for the detection and rating of narrativity. Specifically, we test Llama 3.1 Instruct models with 8B and 70B parameters, using prompt-based classification for each narrative feature.

For each feature, we prompt each model in two separate runs. In the first run, the model is asked to detect the presence of the feature as a binary decision. In the second run, the model is asked to rate the feature on a continuous scale, assigning a value between 0 and 1 for \textit{Story} and between 1 and 5 for story features. This allows us to evaluate LLMs' performance both in identifying narrativity as a binary property and in estimating it as a scalar property. Results are shown in Table~\ref{tab:test_llm_results}. The prompts used for these experiments are shown below:
\begin{itemize}
    \item System Prompt:
    \begin{quote}
        \texttt{``You are a narrative analysis expert.''}
    \end{quote}

    \item User Prompt:
    \begin{itemize}
        \item Detecting presence:
        \begin{quote}
            \texttt{``Determine whether the following text contains a [FEATURE]. Output only a single digit, 0 if the text does not include a [FEATURE] and 1 if the text includes a [FEATURE].}\\
\texttt{[FEATURE] definition:
[FEATURE DEFINITION]}\\
\texttt{Text:
[TEXT]''}
        \end{quote}
        
        \item Rating strength:
        \begin{quote}
            \texttt{``Rate the [FEATURE] strength in the following text as a real number between [LOWEST VALUE] and [HIGHEST VALUE]. Output only the number.}\\
            \texttt{[FEATURE] definition: [FEATURE DEFINITION]}\\
\texttt{Text:
[TEXT]''}
        \end{quote}
    \end{itemize}
\end{itemize}

The results show that, for all features, the fine-tuned encoder-based models outperform both Llama 3.1 Instruct 8B and 70B in detecting and rating narrativity. The performance gap is especially large for the narrative features. In particular, the 70B model performs poorly on \textit{Suspense}, while both Llama models show weak performance on \textit{Agency}. Overall, the 8B model performs better than the 70B model, except for the \textit{Story} and \textit{Event Sequencing} features. No consistent rating error pattern is observed across the two models.

Focusing on LLMs, binary predictions generally achieve higher performance than predictions obtained by binarizing scalar ratings. In contrast, for the encoder-based story models, the opposite pattern is observed (Table~\ref{tab:test_results}): the model trained on soft labels, whose ratings are later binarized, outperforms the model trained on binary hard labels. Agreement between binary predictions and binarized ratings is moderate for both Llama models, with an average Cohen's $\kappa$ of 0.457 for the 8B model and 0.564 for the 70B model.

These results are consistent with prior findings by \citet{bamman2024classificationlargelanguagemodels} showing that Llama 8B and 70B do not outperform encoder-based models on narrativity detection.

\begin{table}[h]
\caption{Performance of the encoder-based models compared to Llama 8B (L.8B) and Llama 70B (L.70B) on the test set, using our best fine-tuned model per feature (DistilBERT(D), BERT(B), RoBERTa(R), and ModernBERT(M)). For \textit{Story}, we use the scalar-based model given its higher performance. We also report the performance of the binarized LLMs scalar ratings (L.8B\(_{binarized}\) \& L.70B\(_{binarized}\)).  Metric ranges are as follows: F1 [0,1] (higher is better), and RMSE [0,1] for \textit{Story} and [0,4] for story elements (lower is better).}
\label{tab:test_llm_results}
\centering
\begin{tabular}{l|c|ccccc|ccc}
\toprule
\multirow{2}{*}{\textbf{Feature}} 

& \multirow{2}{*}{\textbf{Model}} & \multicolumn{5}{c|}{\textbf{F1 $\uparrow$}} & \multicolumn{3}{c}{\textbf{RMSE $\downarrow$}}  \\ \cline{3-10} & &\textbf{\textsc{Argus}} & \textbf{L.8B} & \textbf{L.8B\(_{binarized}\)} & \textbf{L.70B} & \textbf{L.70B\(_{binarized}\)} & \textbf{\textsc{Argus}} & \textbf{L.8B} & \textbf{L.70B}  \\ 
\midrule
Story       &  R & \textbf{0.842} & 0.784 & 0.614 & 0.830 & 0.769 & \textbf{0.194} & 0.323 & 0.291 \\
Suspense     &  B   & \textbf{0.771} & 0.565 & 0.584 & 0.197 & 0.036 &\textbf{0.554} & 1.187 & 1.494 \\
Curiosity    & R & \textbf{ 0.871} & 0.733 & 0.714 & 0.468 & 0.409 &\textbf{0.564} & 1.497 & 1.492 \\
Surprise     & D & \textbf{0.795}  & 0.709 & 0.678 & 0.613 & 0.484 &\textbf{0.612} & 1.716 & 1.340 \\
Agency       &   M    & \textbf{0.714} & 0.493 & 0.522 & 0.486 & 0.534 &\textbf{0.843} & 3.198 & 3.434 \\
Event Sequencing   & D & \textbf{0.765} & 0.596 & 0.536 & 0.725 & 0.656 & \textbf{0.771} & 1.569 & 1.006 \\
\bottomrule
\end{tabular}
\end{table}